# A Method for Planning Given Uncertain and Incomplete Information.

Todd Michael Mansell*
Department of Computer Science
The University of Melbourne
Parkville, Victoria 3052, Australia.
email: tmm@mulga.cs.mu.oz.au.

## Abstract

This paper describes ongoing research into planning in an uncertain environment. In particular, it introduces U-Plan, a planning system that constructs quantitatively ranked plans given an incomplete description of the state of the world. U-Plan uses a Dempster-Shafer interval to characterise uncertain and incomplete information about the state of the world. The planner takes as input what is known about the world, and constructs a number of possible initial states with representations at different abstraction levels. A plan is constructed for the initial state with the greatest support, and this plan is tested to see if it will work for other possible initial states. All, part, or none of the existing plans may be used in the generation of the plans for the remaining possible worlds. Planning takes place in an abstraction hierarchy where strategic decisions are made before tactical decisions. A super-plan is then constructed, based on merging the set of plans and the appropriately timed acquisition of essential knowledge, which is used to decide between plan alternatives. U-Plan usually produces a super-plan in less time than a classical planner would take to produce a set of plans, one for each possible world.

## 1 Introduction

Planning has been a core area of AI research as it explores a key aspect of intelligent activity, the ability to determine a course of action that is likely to achieve a desired goal. Much work has been devoted to understanding various aspects of the planning problem, such as appropriate representations or computationally effective search strategies. Generally, the problem has been approached from two quite distinct perspectives. On the one hand, the problem has been presented as one of finding a 'provably correct' plan for achieving a stated goal in a static world for which there is a complete description. On the other hand, the problem has been presented as one of determining the next action to take in a dynamic world that defies complete description and for which the 'correct' action is determined by matching characteristics of the world to preconditions of predetermined responses. We wish to plan in circumstances that are intermediate to these two extremes. We want to determine a plan that may (will) include alternative actions to achieve, at least to some degree, a desired goal in a world for which an incomplete description exists.

The world in which we wish to plan lacks a complete description and consequently alternative plans may need to be constructed when the description is insufficient to eliminate alternatives. We assume (require) that additional information about the world can be acquired by knowledge acquisition activities. These activities are likely to incur a cost. We wish to find a plan that attains the desired goal by acquiring only necessary supplementary information. In addition, we only want to acquire supplementary information at the point in the plan when that information is needed to select among alternative actions.

The worlds in which we wish to apply our techniques are dynamic. Other agents in our world may change its state in unpredictable ways. However, we assume that our incomplete description of the world is static and we plan and select our first action based on this description. We will reassess the world after executing our first action and determine a new plan, if necessary, to attain the desired goal. The notion we adopt is that the world at its most detailed may change during planning, but the more abstract concepts are unlikely to change significantly. Hence, the tactical detail of a plan may need modifying, but the plan strategy will still be relevant.

Goals, in many applications, are not precise requirements. Many general goals can be fulfilled to various degrees by achieving alternative subgoals.

*Present address Materials Research Laboratory, PO Box 50, Ascot Vale, 3032, Melbourne, Australia.

However, not all subgoals are equally likely to be achieved. We adopt an approach to planning by determining a course of action that is likely to maximise the expected fulfilment of our goal. Consequently, our plans are not exhaustive. They do not elaborate all the alternative actions required in all possible worlds. Rather, they specify alternative actions that are likely to maximise the expected fulfilment of our goal in the possible worlds that are consistent with our partial description of the world.

The problem of planning under uncertainty is currently being addressed by a number of researchers. Lowrance and Strat[Low90] have developed a system that constructs plans given incomplete and uncertain information about the initial state of the world, and operators with probabilistic outcomes. A plan is constructed for the each state using SIPE[Wil88], and a likelihood that the plan will achieve the goal calculated using GISTER[Low91], an uncertain reasoning system.

Bonissone and Dutta[Bon90] present an approach for planning in an environment that is dynamic and where the effects of various actions on the world is uncertain. T-norms[Bon87] are used to represent uncertainty in both information and system rules in a planning hierarchy.

SUDO-Planner[Wel90] uses tradeoff formulation to separate significant decisions from trivial choices. SUDO-Planner constructs plans using actions with uncertain effects and allows for partial goal satisfaction. In other work, Dean[Dea90] examines how one can make predictions in an uncertain environment given temporal constraints.

## 1.1 U-Plan

A major problem when planning given incomplete and uncertain information about the environment is that it is not possible to construct one initial state that precisely and unambiguously represents the world. U-Plan uses a possible worlds representation, where the available initial information is used to construct every possible initial state of the world. Associated with each possible world is a numerical measure of belief specifying the degree to which the evidence supports each possible world as the one that represents the true state of the world.

A hierarchical approach to planning is used as it significantly reduces the search space by first planning at abstract levels, and then expanding these abstract plans into more detailed plans. At the highest abstraction level strategic decisions are made, while at the lowest levels of abstraction, tactical decisions about how best to implement the strategy, are made. In support of hierarchical planning, each possible world is described at a number of predefined abstraction levels, allowing decisions to be made using a state representation at an equivalently detailed level of abstraction.

Hierarchical planning selects an overall strategy before becoming involved with the tactical detail. U-Plan utilises a set of (predefined) goal reduction operators that encode how a planning goal is reduced by the operator's application. What results is a planning hierarchy tree where the goals are broken up into subgoals by the goal reduction operators. This allows us to first make the strategic decisions, which then guides all other decisions down to the tactical implementation of the subgoals. The reduction operators are expressed at various levels of abstraction. The planning algorithm uses the same abstraction level for both operator and possible world. A measure of expected fulfilment is used when selecting which operator to apply next.

U-Plan constructs a plan for one possible world at a time, the first plan being constructed for the possible world with the greatest likelihood of representing the true world. Before subsequent plans are constructed, the suitability of reapplying an existing plan to this new possible world is assessed. Associated with each plan is the possible worlds they work for. If a plan partially works for another possible world (e.g. the strategy works but some of the detail is different), then that part of the plan is used for this possible world, and planning continues from where the plan failed. When a plan exists for every possible world, the operator order of all the plans is combined to obtain a single planning tree that branches when the operator execution order differs. At this point the ability to acquire additional knowledge is used. At each branch, a knowledge acquisition operator can be inserted to determine which action in the planning tree to carry out next.

This planning algorithm is presently being tested using an air combat domain. In this example, the goal is to successfully attack a target aircraft, given only partial information about the target location, type, and status. A number of strategies exist on how the attack should be carried out. Each strategy uses a different method of attack, and therefore has a different probability of success, and a different degree to which it fulfils the goal.

In order to assess the performance of U-Plan a control planning system, C-Plan, has been developed. C-Plan is a linear, hierarchical planner that uses a decision theoretic operator selection process. The main differences between the two planning algorithms are C-Plan does not attempt to reapply plans, merge identical plans, or acquire knowledge. The majority of C-Plan's code is borrowed from U-Plan to ensure a comparison between planning algorithms (and not the efficiency of the code).

Central to planning using U-Plan is: the set of states are represented at several abstraction levels; the selection of reduction operators is not purely state based but dependent on a calculation of expected fulfilment; the system will attempt to acquire additional knowledge when it is advantageous to do so, and an attempt is made to apply an existing plan to more than one initial state.

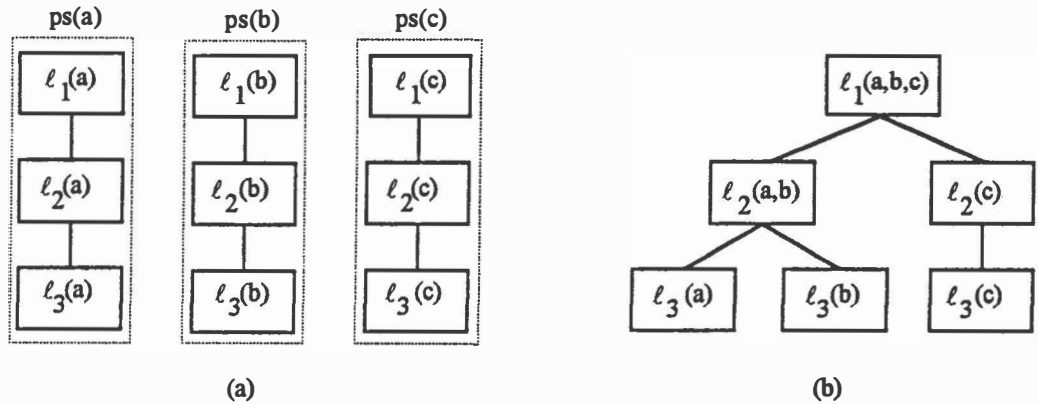

Figure 1: (a) Depicts how 3 sample P-states have representations at 3 abstraction levels. $\ell_n(x)$ represents P-state, x, at abstraction level, n. (b) An example of how 3 initial P-states may be grouped in tree form.

## 2 State Representation

Classical planning systems assume the availability of complete knowledge at the time of planning. Unfortunately, in real environments, an agent must often deal with incomplete information, (due to, for example, the sensory limitations of knowledge sources).

### 2.1 P-States

When an incomplete model of the world is all that is available, a set of initial states can be used to describe the alternative environments. U-Plan employs a set of initial possible states (P-states) to describe what might be true of the world. A P-state, ps(a), is a complete description of one possible world using propositional statements. Each P-state is described hierarchically with n levels of abstraction, $(ps(a)=\{\ell_1(a) \ldots \ell_n(a)\})$ where n is domain dependent and selected during knowledge engineering (fig. 1(a)). The level $\ell_i(a)$ is a complete description of a world at the $i^{th}$ level. The highest level of abstraction gives a coarse description of the state of the world. The lowest level gives a detailed view of the world. Intermediate levels provide the description required to make a smooth transition between both extremes.

Information sources provide U-Plan with a set of propositional statements, that represent distinct aspects of the domain. Each propositional statement has associated with it measure of certainty (U-Plan uses a Dempster-Shafer mass distribution for reasons discussed in section 2.3). The propositional statements are then mapped to the lowest level of abstraction where they are used to generate a set of detailed P-states (For example, in figure 1(a) the initial information is used to construct $\{\ell_3(a),\ell_3(b),\ell_3(c)\}$, the set of P-states described at the lowest level of abstraction).

Mapping functions (defined at knowledge engineering time and domain dependent) are then used to construct a representation of the detailed state at the next level of abstraction. This process continues till each state is described at every level of abstraction. A P-state is one possible world with a description at every abstraction level (e.g., P-state, ps(a), is also represented by $\{\ell_3(a),\ell_2(a),\ell_1(a)\}$, giving a description of the same possible world at differing levels of abstraction).

### 2.2 P-State Grouping

U-Plan groups together equivalent initial P-states according to their hierarchical levels, ie., the P-states with the same state description at a particular abstraction level are grouped together.

Figure 1(b) demonstrates how initial P-states may be grouped in tree form. In this example the set of P-states from figure 1(a) are used. At the lowest level of abstraction the set of possible worlds are distinct, represented as the leaf nodes of the tree, $\{\ell_3(a),\ell_3(b),\ell_3(c)\}$. Let us assume, when viewing the world in a more coarse light, ie. at a higher level of abstraction, $\ell_2(a)$ and $\ell_2(b)$ are identical. In this case they would be grouped together to give $\ell_2(a,b)$. At the highest level of abstraction $\ell_1(a,b)$ and $\ell_1(c)$ might also be identical resulting in the state $\ell_1(a,b,c)$.

### 2.3 P-State Ranking

Information acquired in a real-world situation provides evidence about the possible states of the world. This information is typically uncertain and incomplete. Dempster-Shafer (D-S) Theory[Sha76] is one way of handling such evidence, using an interval to explicitly capture what is known as well as what is not known (i.e., uncertainty). This is achieved by allowing belief to be associated directly with a disjunction of events. The D-S belief interval was chosen for its ability to capture the essence of second order statistics using first order assessments. U-Plan is not dependent on this representation, probability theory or fuzzy logic could also be implemented with minimal effect to performance.

Fundamental to D-S Theory is the frame of discernment (or frame), $\Theta_A$, the set of mutually exclusive, exhaustive propositional statements representing what is known of the world. For example, frame A might represent the various headings a target aircraft may take. Propositional statements are represented by disjunctions of elements of the frame ($\Theta_A$).

U-Plan represents each piece of information about the environment as a propositional statement within a frame of discernment. A unit of belief is distributed over the frame associating an accuracy to the proposition. This is called the mass distribution, where the mass assigned to each proposition is between 0 and 1.

Belief in a proposition $A_j$ is represented by the evidential interval, $[Spt(A_j), Pls(A_j)]$, where

$$Spt(A_j) = \sum_{A_i \subseteq A_j} m_A(A_i),$$

and $Pls(A_j) = 1 - Spt(\neg A_j)$. Here, $Spt(A_j)$ represents the degree to which the evidence supports the proposition, and $Pls(A_j)$ represents the plausibility of $A_j$ (i.e., the degree to which the evidence fails to refute the proposition). A measure of support and plausibility is calculated for each initial P-state at every level of abstraction based on the mass distributions of the initial evidence.

Additional aspects of the domain can be incorporated into the system by the inclusion of new frames of discernment. Compatibility relations[Low91, Sha76] are then used to describe which elements from two frames can be true simultaneously; allowing propositional statements to be addressed jointly. For example, $\Theta_B$ might represent altitude of a target aircraft, and the compatibility relation maps $\Theta_A$ and $\Theta_B$ to a new frame $\Theta_{A,B}$ which (in this case) is the cross-product of the two sets.

Compatibility relations are used by U-Plan to specify the interrelationships between a piece of information at one level of abstraction, and other levels of abstraction. The restriction is that a compatibility relation can only link one level to the levels directly above or below it. To ensure that each level is a complete representation of the possible world, a compatibility relation must exist for every element of the frame.

### 2.4 P-state Selection

The selection of the initial P-state to begin planning involves choosing the P-state with greatest support[1] at the highest level of abstraction, (for example $\ell_1(a,b,c)$).

The node in the P-state tree that is a child of this initial P-state with the greatest support is then selected (e.g. $\ell_2(a,b)$ or $\ell_2(c)$). This selection process continues from highest to lowest level of abstraction. The result is an initial P-state with a description at all levels of abstraction.

The P-states are chosen in this manner in an attempt to allow the possible world with the greatest support to be planned first. This does not guarantee the plan will have the greatest support when planning is complete, or that the best plan will be constructed first. The usefulness of this strategy becomes apparent in section 6 when attempting to use all or part of previously constructed plans during planning for other P-states. The effectiveness of this approach relies on a suitable representation of the domain and the reduction operators.

## 3 Reduction Operator

Planning operators represent actions that the system may perform in the given domain. The role of an action is to change the state of the world, the aim of an operator is to represent how applying that action will change the system's view of the state of the world. U-Plan uses reduction operators to give alternative methods for achieving the goal at a lower level of abstraction, or at the tactical level it describes the direct effects of an action on the P-state. These are ABSTRIPS-like operators where the closed world assumption is implemented, and hierarchical planning used.

Defined for each operator is the:

• Abstraction Level: designating the P-state's level of abstraction it operates on.

• Necessary Preconditions: the wffs that must be true in the P-state before the operator can be applied. The system will make no attempt to make these precondition true.

• Satisfiable Preconditions: represent conditions that must be true of the world before the operator can be applied. U-Plan will attempt to find an operator/operators that satisfy these preconditions.

• Plot: provides step-by-step instructions on how to perform the action represented by the operator. This includes a description of the goal reduction operators that are applied at the next level of abstraction, and it's fulfilment, measuring the degree to which it achieves the goal of the present operator. Or, at the lowest level of abstraction, how the operator changes the P-state.

• Probability[2]: a function for calculating the probability of the reduction operator succeeding given the current P-state. The availability of such a function is domain

---

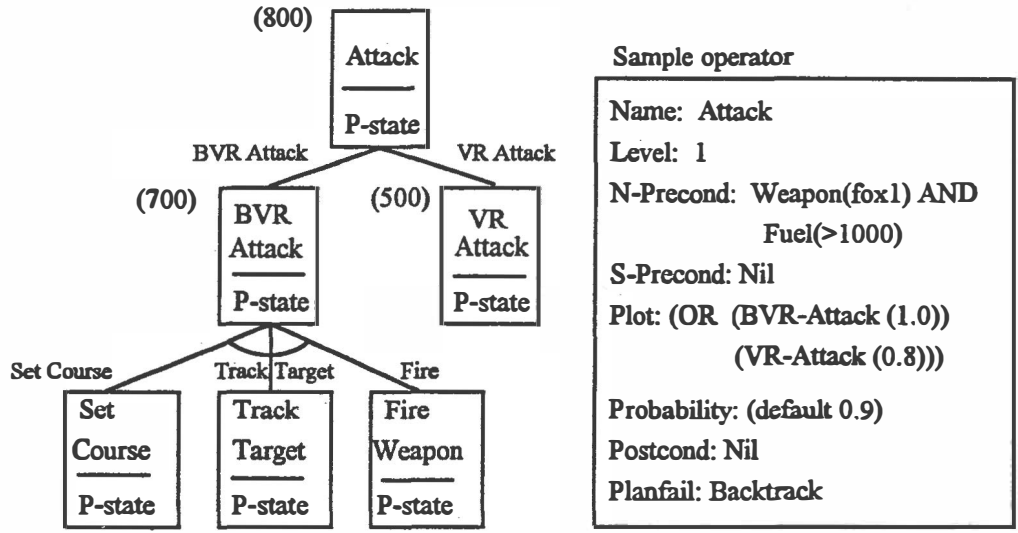

Figure 2: Part of the strategy hierarchy for a simplified air combat example, and a simplified example of an operator.

specific and non-trivial. In the example introduced here (figure 1) the function is obtained empirically (based on historical data).

- Postconditions: what the operator achieves.

- Planfail: what to do if the operator fails at planning stage.

The probability of success does not provide sufficient information to select a reduction operator as it does not take into account the goals of the system. It is for this reason that associated with each reduction operator listed in the plot of a parent reduction operator is a measure of fulfilment, representing the degree to which the reduction operator achieves the goal of the parent. The reduction operator's expected fulfilment (representing the value of an action satisfying the goal of its parent given the current state of the world) is calculated during planning and utilised in reduction operator selection.

### 3.1 Causal Theory

U-Plan uses a deductive causal theory[Wil88, Ped88] to deduce the context dependent effects of applying a reduction operator to a P-state. The effects that are deduced are considered to be side effects, where those that are introduced directly by the reduction operator are the direct effects. The use of deduced effects simplifies the description of the operators by removing the need for extensive add and delete lists. After the application of each reduction operator a set of triggers are used to determine if the world has been changed in such a way that the deductive rules need be applied. If so, the deductive causal theory is used to change the P-state to be consistent with all the effects of an action. The side effects of applying any reduction operator are recorded in the planning hierarchy tree.

## 4 Representation of Plans

U-Plan does not construct a state-based search tree, but constructs a strategy hierarchy which is a decision tree like structure, where the nodes in the hierarchy represent a continuous transition of actions from the strategic (at the root node) to the tactical (at the leaf nodes). The nodes closest to the root node are highest in strategic intent representing not only a decision at a high level of abstraction, but the direction the planning will take. The nodes closest to the leaf nodes have maximum detail representing task and action sequences.

The strategy hierarchy can be represented as an AND/OR search tree, the root node representing the strategic goal of the system, and the leaf nodes representing the tactical details of how the goal is to be achieved. Each node in the tree is a subgoal node representing the current goal and P-state, and certain pairs of nodes are connected by arcs representing the application of a reduction operator that produces this subgoal node. For example, figure 2 shows part of the strategy hierarchy for a simplified air combat domain, where the goal is to engage an opponent (not shown), and one way to do this is to Attack. The strategy hierarchy shows that there are two reduction operators that can be applied to achieve an attack, the BVR Attack (or Beyond Visual Range attack) and the VR Attack (or Visual Range attack). These operators redefine the aims of the system (represented by the nodes in the tree) to the BVR attack subgoal, and the VR attack subgoal respectively.

When selecting a reduction operator, the expected fulfilment of each is calculated to aid in the decision. The calculation is based on the reduction operator and the P-state attributed to the subgoal it produces. (In figure 2 the expected fulfilment appears in brackets to the left of each subgoal node.) Each single subgoal node represents part of a plan to achieve the subgoal at the next highest level of abstraction.

## 5 The Basic Planning Algorithm

Many classical planning systems use a state-based search strategy to solve planning problems. To find a solution one applies operators to a state description until an expression describing the goal state is found. U-Plan employs decision theoretic techniques in an abstraction hierarchy for operator selection.

This planning algorithm. constructs linear plans which describe a sequence of actions that, when applied in a particular P-state, have a given probability of producing the goal state.

Initially, U-Plan is given a goal to achieve. This goal function is placed as the PLANHEAD node of a plan tree. The set of reduction operators that satisfy the goal node are obtained from the plot of the goal node. These reduction operators are added to the planning hierarchy, and the subgoal they produce are constructed. The expected fulfilment for each of the subgoals is then calculated, by multiplying the operators fulfilment with it's probability . The fulfilment is determined from the degree to which the operator achieves the goal of the parent operator; and the probability is gained from the probability function associated with the operator.

The planning algorithm for U-Plan uses the expected fulfilment values to select which reduction operator in the strategy hierarchy is to be expanded next. If the necessary preconditions of this reduction operator are true in the active P-state, then the reduction operator is selected, else the plan fail is applied (this usually involves backtracking). When a reduction operator that satisfies the necessary preconditions has been found, the satisfiable preconditions are tested. If any of these are not true, U-Plan can attempt to satisfy them using reduction operators of equal or lower abstraction. If these preconditions are not satisfied, the node is rejected, then the planfail procedure is adopted.

Once both sets of preconditions of the reduction operator can be shown to be true in the active P-state, the plot can be applied. The plot represents the effects the reduction operator has on the state of the world, and the subgoals that may be used to achieve this subgoal. When applying the plot, the next level of the strategy hierarchy is exposed, and again the subgoal with the highest expected fulfilment is selected to be expanded next. At the lowest level of abstraction the reduction operator specifies how the P-state is changed.

This sequence of selecting and applying reduction operators continues from the highest level of abstraction to the lowest. In our air combat example, an Attack can be achieve by applying a BVR Attack or a VR Attack. The expected fulfilment of a BVR attack (in parenthesis beside the operator) is greater than that of a VR attack, hence it is pursued.

When constructing a strategy hierarchy it is possible that as a plan's detail is filled out it becomes less likely to succeed. This is because our strategic decisions are based on information at a coarser level of abstraction. This makes it important to review earlier decisions while planning. After the application of a group of reduction operators U-Plan compares the expected fulfilment of the current subgoals, with those of previous subgoals, and determines if they fall below the previous values less an offset. Including an offset is an iterative deepening strategy[3]. The offset value will depend on the difference in abstraction level of the subgoals. It is expected that as the system uses lower. level information the expected fulfilment of the plan will decrease. This offset value helps avoid the problem of the system jumping around from branch to branch in the strategy hierarchy.

## 6 Plan Reapplication

U-Plan applies plan reapplication in an attempt to determine if a plan generated for one initial P-state can be adopted for another initial P-state. The desired result being fewer plans than the number of initial P-states. This is accomplished by first generating a plan for the initial P-state with the greatest support, then apply it to the next highest supported state to see if it achieves the goal function. Similarly, the rest of the initial P-states, with a support and plausibility over some threshold, are selected to see if the plan works in their situation. The initial P-states the plan works for are recorded and this information used in plan selection.

A plan is reapplicable if all the reduction operators in the plan (that are not redundant) have their preconditions met under the new initial P-state, and when applied result in the goal state being achieved. That is, each reduction operator in the linear plan is applied to the new P-state in order, and if all operators succeed and the goal state is reached, the plan has been successfully re-applied.

If a plan, during reapplication, fails due to the unsuccessful application of an operator, that plan is not entirely discarded. U-Plan will attempt to use the part of the plan that was successful and planning continues from the point where the plan failed. The desire is to construct plans with the same or similar strategies by reusing, at least part of, the plan at the high level of abstraction.

---

[3]A number of iterative deepening strategies exist that can be applied to this problem of selecting a suitable offset between different abstraction levels.

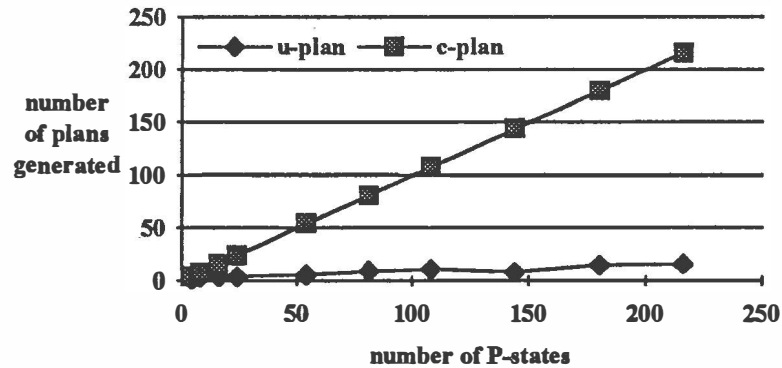

Figure 3: Comparison of the number of plans generated by U-Plan and C-Plan
in the air combat domain.

The case may arise when more than one plan partially works for a new initial P-state. A number of options are available as to which part plan to continue planning from; one is to select the plan with the greatest expected fulfilment; another is to select the plan with the greatest support. Both these options are appealing for different reasons (depending on the domain and type of plan desired) and are available for selection at knowledge engineering.

## 7  Super-Plans

Once plans exists for all the P-states, with plausibility above some threshold, a single super-plan is constructed. This is achieved by merging set of plans constructed for the set of initial P-states with the aid of knowledge acquisition operators. To simplify the description, let us assume there are two plans. The super-plan begins with the initial actions common to both plans (if any), these actions could be applied before any information is acquired. When the sequence of actions of the plans differ a branch in the super-plan is added and the remainder of each plan constitutes each branch of the super-plan. At each branch in the super-plan a knowledge acquisition operator is added, attaining the information required to select which action in the super-plan to apply next.

The case may arise that the required information is not available and an action must be selected, based on the support for each branch of the super-plan. The mass one attributes to each plan is given by:

$$m(p_i) = \sum_{\Psi \subseteq \Phi} m(\Psi)$$

where $p_i$ is the plan, and $\Phi$ is the set of all initial P-states that use $p_i$ as their plan. Simply stated, the mass of each plan is equal to the sum of the mass of the initial P-states that use the plan; this includes the mass attributed to the disjunction of any of the initial P-states in the set. The mass associated with each branch of the super-plan is therefore equal to the sum of the masses of the plans included in that branch.

For example, if the plan $p_1$ works for initial P-states ps(a) and ps(b), then the mass attributed to $p_1$ is equal to the sum of mass attributed to ps(a), ps(b), and {ps(a) ∨ ps(b)}. If this is the only plan included in the branch of the super-plan and it has a greater mass than the other branches, then this branch is selected.

When deciding whether to acquire information it is important to evaluate the cost as well as the benefit. This cost may be of the form of the time it takes, the resources it uses, the direct effect it has on the environment, or the information it also gives to an opponent. A trade-off exists between when to plan to acquire information, based on the cost to do so, and when to select a branch, based on belief that one of the P-states represented by the plan is the true state of the world. For example, do you turn on your radar to find an opponents altitude and give away your location when you are 90% sure of its altitude. This trade-off is presently implemented by U-Plan, using a heuristic based on belief and cost.

## 8  Results

In order to assess the operation of U-Plan a control planning system, C-Plan, has been constructed. C-Plan is a hierarchical planner that uses the same decision theoretic operator selection process as U-Plan, but does not attempt to reapply plans, merge plans, or acquire knowledge.

When planning given a set of possible worlds, C-Plan will construct a plan for every initial P-state. The number of plans generated by U-Plan depends on the domain. In the air combat domain, U-Plan produces substantially fewer plans than C-Plan, see fig. 3. The reduction in the number of plans here is aided by the type of the domain which, by its nature, sustains a number of possible planning solutions. In a worst case domain, ie. one where one unique plan exists for each possible world, U-Plan produces one plan for each initial P-state.

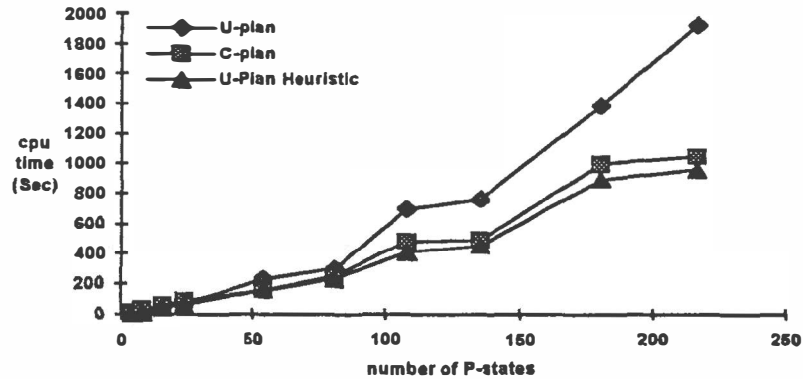

Figure 4: Comparison of the amount of cpu time used by U-Plan, C-Plan and U-Plan(heuristic) for various numbers of initial P-states on a Symbolics 3645.

The process of reapplying plans can be costly. The more plans U-Plan generates the larger the set of possible plans to be reapplied. The effect of this overhead can be alleviated by the implementation of some simple heuristics. In the air combat domain heuristics can be used to quickly evaluate the suitability of a plan to the current P-state. Figure 4 plots the amount of cpu time spent running C-Plan, U-Plan and a version of U-Plan that uses heuristics during plan reapplication. This demonstrates that, when heuristics are available, U-Plan constructs fewer plans and intends to acquire the knowledge to differentiate between them with no time penalty over planning for every possible state. When heuristics are not available, the reapplication of plans becomes the overriding factor in the amount of time spent planning.

The heuristics used in the alternate version of U-Plan involve running a rough test to determine whether the plan *is likely to fail for a given P-state*. The test *itself* involves looking for known predicate-operator combinations in the plan that are known to be incongruous. In the air combat example used here, certain predicates that make up the P-states are incompatible with particular strategies in plans. This particular heuristic test only guarantees to determine which plans will not work for certain P-states. If the heuristic test is passed, a rigorous examination is plan is carried out. However, as the standard U-Plan in figure 4 spent between 50-90% of its time searching for existing solution plan, the largest savings are likely to be made here.

Figure 4 shows U-Plan tends to closely shadow the cpu times of the control planner. However, our suspicion is that, for a problem with a large number of P-states the system overheads associated with managing the swap space for the extensive set of plans generated by C-Plan, would see an increased benefit in system time for U-Plan.

In the air combat example, U-Plan produces plans that, mostly, reuse a small number of strategies. In many of the scenarios the plans produced have the first few detailed actions in common. At plan execution, these actions are applied before any information is required to be obtained. At this point, the environment is re-assessed and a judgment (bases on the significance of the changes) made whether to continue the execution of the super-plan, or to begin replanning using part or none of the existing plans.

## 9 Conclusion

This paper outlines the ideas behind U-Plan, a system for planning given an uncertain environment and incomplete information. The system represents the incomplete and uncertain description of the environment using a set of possible worlds. Each of these possible worlds contains a representation of the world at a number of abstraction levels. This enables the planning system to make high level (strategic) decisions, based on a high level representation of the world. As more abstract world representations generally encompass a number of low level possible worlds, the high level planning can develop a strategy for a number of possible worlds. This allows the system to commit more readily to a plan strategy, when given a number of possible states of the world.

The planning system's ability to fit part or all of an existing plan to a number of possible worlds, has the potential to produces fewer plans than one for each possible world. The computational cost of planning in this manner depends on the domain of application, and the relative number of plans produced. Any computational savings over a decision theoretic hierarchical planner constructing plans for each possible world, relies on two properties. U-Plan constructing fewer plans than one per P-state; and the application of heuristics to quickly and accurately determine whether attempting to reapply a plan to another P-state is worthwhile. As U-Plan's speed depends on producing a manageable number of possible plans, the system is targeted towards domains where a number of possible plans exist for the possible worlds. Hence, planning is based on constructing a superior plan that achieves the system's goals given the available information. Conversely, a domain that required a unique plan for every possible world would be more computationally costly to plan for under U-Plan.

The inclusion of the system's ability to plan to acquire information when beneficial to do so, enhances U-Plan's operation. What results is a plan tree that exploits common action sequences, and provides the mechanism to select which branch to pursue in the plan tree. This yields a more favourable result than producing a new plan for every possible world.

U-Plan is intended for use within domains where decisions must be made before full knowledge is available. In particular, emergency type domains where actions should proceed while information is being collected; or complex environments where complete information is unobtainable. U-plan is currently being applied to a fire hazard action response domain, in which a super-plan is constructed to combat industrial fires.

## Acknowledgments

I would like to thank Grahame Smith of the Australian Artificial Intelligence Institute, and Elizabeth Sonenberg of The University of Melbourne for their many insightful comments on this work.

This work was completed at The University of Melbourne while the author was employed as a postgraduate research fellow by the Materials Research Laboratory of the Australian Defence Science and Technology Organisation..